\pdfoutput=1

\documentclass[11pt]{article}

\usepackage[]{emnlp2022}

\usepackage{times}
\usepackage{latexsym}
\usepackage{listings}
\usepackage{amsmath}
\usepackage{multirow}
\usepackage{adjustbox}
\usepackage{subcaption}
\usepackage{graphics}
\usepackage{graphicx}

\usepackage[T1]{fontenc}

\usepackage[utf8]{inputenc}

\usepackage{microtype}
\usepackage{graphicx}
\usepackage{inconsolata}
\usepackage{lipsum}

 \newcommand{\Smara}[1]{\textcolor{blue}{\bf \small [ #1 --Smara]}}

\usepackage{xspace}
\newcommand{\flute}{\textsc{flute}\xspace}

\makeatletter
\newcommand{\printfnsymbol}[1]{%
  \textsuperscript{\@fnsymbol{#1}}%
}
\title{FLUTE: Figurative Language Understanding through Textual Explanations}

\author{Tuhin Chakrabarty$^1$~~~~~Arkady Saakyan$^{1}$~~~~~Debanjan Ghosh$^{2}$~~~~~Smaranda Muresan $^{1}$ \\
 $^1$Department of Computer Science, Columbia University\\  
 $^2$Educational Testing Service\\
 {\tt\small tuhin.chakr@cs.columbia.edu, a.saakyan@columbia.edu, dghosh@ets.org, smara@cs.columbia.edu} \\
}

\begin{document}
\maketitle
\begin{abstract}

Figurative language understanding has been recently framed as a recognizing textual entailment (RTE) task (a.k.a. natural language inference, or NLI). However, similar to classical RTE/NLI datasets, the current benchmarks suffer from spurious correlations and annotation artifacts. To tackle this problem, work on NLI has built explanation-based datasets such as e-SNLI, allowing us to probe whether language models are right for the right reasons.  Yet no such data exists for figurative language, making it harder to assess genuine understanding of such expressions. To address this issue, we release \flute, a dataset of 9,000 figurative NLI instances with explanations, spanning four categories: Sarcasm, Simile, Metaphor, and Idioms. We collect the data through a model-in-the-loop framework based on GPT-3, crowd workers, and expert annotators. We show how utilizing GPT-3 in conjunction with human annotators (novices and experts) can aid in scaling up the creation of datasets even for such complex linguistic phenomena as figurative language. The baseline performance of the T5 model fine-tuned on \flute shows that our dataset can bring us a step closer to developing models that understand figurative language through textual explanations.
\end{abstract}

\section{Introduction} \label{section:intro}
Figurative language such as metaphors, similes or sarcasm plays an important role in enriching human communication, allowing us to express complex ideas and emotions in an implicit way \cite{roberts1994people,fussell1998figurative}. However, understanding figurative language still remains a bottleneck for natural language processing \cite{shutova2011computational}. Recently \citet{jhamtani-etal-2021-investigating} show that when faced with dialog contexts consisting of figurative language,some models show very large drops in performance compared to contexts without figurative
language. Despite the fact that Transformer-based pre-trained language models (LMs) get even larger \cite{brown2020language,raffel2020exploring}, they are still unable to comprehend the physical world, cultural knowledge, or social context in which figurative language is embedded \cite{bisk-etal-2020-experience}.

In recent years, there have been several benchmarks dedicated to figurative language understanding, which generally frame ``understanding'' as a recognizing textual entailment (a.k.a natural language inference (NLI)) task --- deciding whether one sentence (premise) entails/contradicts another (hypothesis) \cite{chakrabarty-etal-2021-figurative,stowe-etal-2022-impli,bigbench}. However, similar to general NLI datasets, these benchmarks suffer from spurious correlations and annotation artifacts \cite{mccoy-etal-2019-right,poliak2018hypothesis}. These can allow large language models (LLMs) to achieve near human-level performance on in-domain test sets, yet turn brittle when evaluated against out-of-domain or adversarial examples \cite{glockner-etal-2018-breaking,ribeiro2016should,ribeiro-etal-2020-beyond}. To tackle these problems, research in NLI has argued that it is not enough to correctly predict the entail/contradict labels, but also to explain the decision using natural language explanations that are comprehensible to an end-user assessing model's reliability  \cite{NEURIPS2018_4c7a167b,majumder2021rationale,wiegreffe2021reframing}, leading to novel datasets such as e-SNLI \cite{NEURIPS2018_4c7a167b}. Yet, there is no such dataset for figurative language, hindering our ability to assess LLMs' genuine understanding of figurative language.

\begin{table*}[t]
    \centering
    \includegraphics[width=\textwidth]{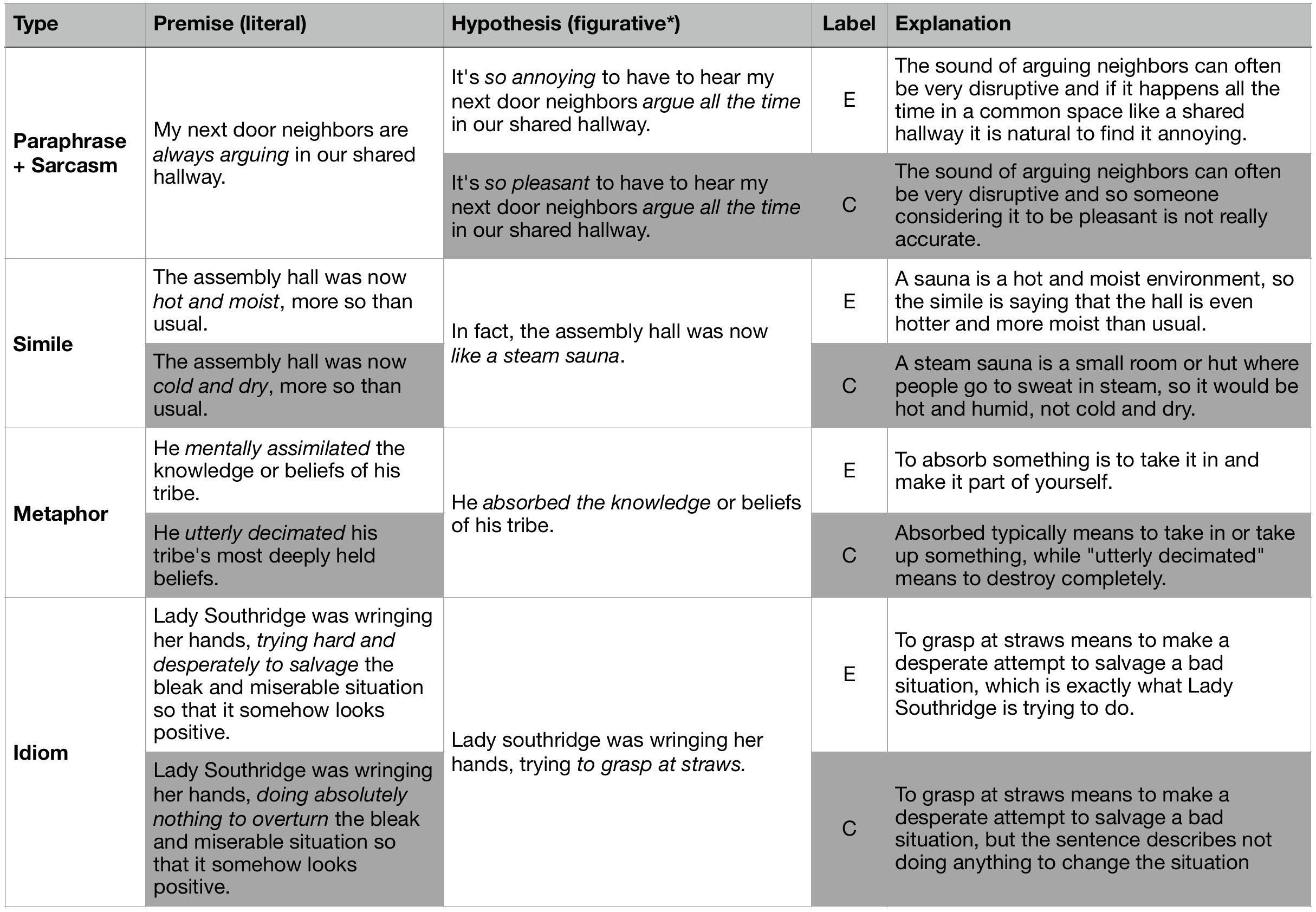}
    \caption{\flute examples of figurative text (hypothesis) and their respective literal entailment(E) and contradiction (C) premises, along with the associated explanations. * For simile, metaphor, and idiom, figurative examples are the hypothesis whereas for sarcasm, we have both figurative and literal hypotheses (see Section \ref{sec:data}).}
    \label{table:data}
\end{table*}

In this paper, we make several contributions towards the goal of building models and assessing their ability to understand figurative language:
\begin{itemize}
    \item {\bf \flute: a new benchmark for figurative language understanding through textual explanations}. FLUTE contains 9,000 high-quality <literal, figurative> sentence pairs with entail/contradict labels and the associated explanations. The benchmark spans four types of figurative language: sarcasm, simile, metaphor, and idiom. Table \ref{table:data} shows examples from our dataset. A noteworthy property of \flute is that both the entailment/contradiction labels and the explanations are w.r.t the figurative language expression (i.e., metaphor, simile, idiom) rather than other parts of the sentence.  
    \item {\bf A scalable model-in-the-loop approach for building \flute.}  Model-in-the-loop approaches (i.e., GPT-3 \cite{brown2020language} and crowdsourcing) have been recently proposed to generate NLI datasets, as well as free-form textual explanations (a.k.a natural language explanations \cite{NEURIPS2018_4c7a167b}) for model decisions \cite{liu2022wanli, wiegreffe2021reframing}. For figurative language, \newcite{ghosh-etal-2020-interpreting} has shown that crowdworkers are mostly good at performing minimum edits to generate a literal sentence from a sarcastic one (e.g., using negation or antonyms), which can lead to trivial examples easily classified by LLMs \cite{chakrabarty-etal-2021-figurative}. Thus, for building \flute, we leverage the power of GPT-3 to generate diverse and high quality literal text (paraphrases/contradictions and/or explanations) using few-shot prompting, coupled with minimal human involvement (e.g., crowdworkers to minimally edit a literal sentence to make it sarcastic and experts for judging and minimally editing GPT-3 output to ensure quality control) (Section \ref{sec:data}). 
    \item{\bf Comprehensive set of experiments 
    to assess \flute's usefulness towards building models that understand figurative language.} We propose a setup inspired by instruction-based learning \cite{mishra2021cross,sanh2021multitask,wei2021finetuned} and train a T5 \cite{raffel2019exploring} model to jointly predict the label (entail/contradict) and explanation. We train two variants: T5 trained on e-SNLI dataset \cite{NEURIPS2018_4c7a167b} and T5 trained on \flute. We evaluate our model on the  \flute test set (Section \ref{section:eval}). We propose extensive automatic and human evaluation experiments to assess model understanding through explanations (Section 
\ref{section:eval}). We show that the model trained on \flute produce higher quality explanations compared to model trained on e-SNLI (Section \ref{section:result}). 
\end{itemize}

Our code and experimental setup is available at \footnote{\url{https://github.com/tuhinjubcse/model-in-the-loop-fig-lang}}.Our data can be accessed at \footnote{\url{https://huggingface.co/datasets/ColumbiaNLP/FLUTE}}

\section{Model-in-the-loop for building \flute} \label{sec:data}
\flute consists of pairs of premises (literal sentences) and hypotheses (figurative sentences)\footnote{Given sarcasm is the opposite of the literal meaning, we would only have contradictions in the dataset, thus we also generate a literal hypothesis that entails the literal premise.}, with the corresponding entailment or contradiction labels (NLI instances), along with explanations for each instance (Table \ref{table:data}). We describe the model-in-the-loop methods for creating premise-hypothesis pairs for each type of figurative language (Section \ref{section:NLI}) and the associated explanations (Section \ref{section:NLE}).

\subsection{\flute: Premise-Hypothesis Pair Creation} \label{section:NLI}

\begin{figure*}[t]
 \centering
 \begin{subfigure}[b]{\textwidth}
  \centering
  \includegraphics[width=\textwidth]{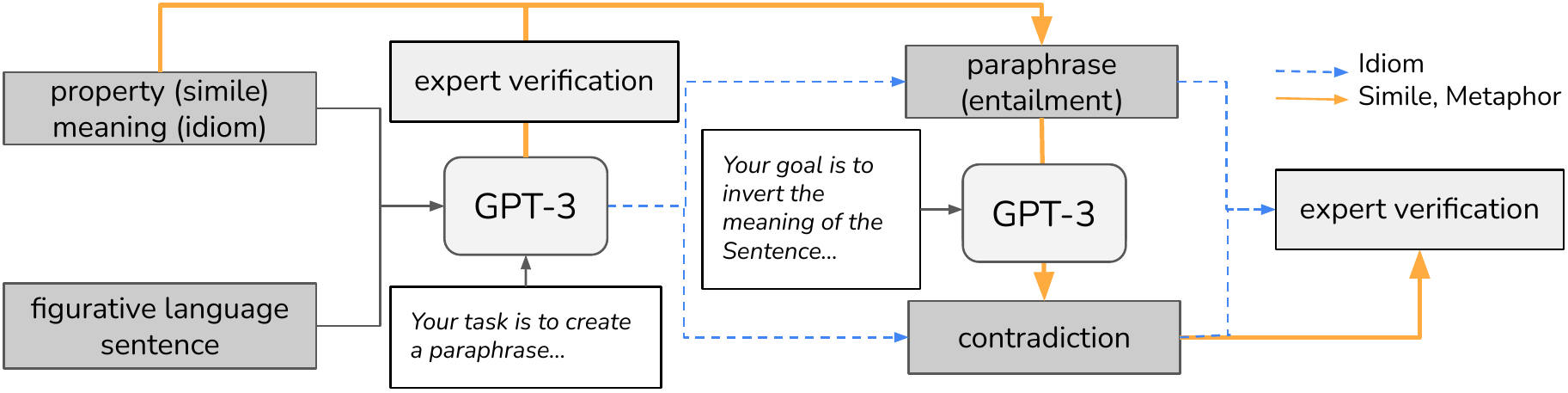}
  \caption{Model in the Loop for \flute: Simile, Metaphor, Idiom Data}
  \label{fig:architecture_allfig}
 \end{subfigure}
\vskip\baselineskip
 \begin{subfigure}[b]{5.3in}  
  \centering 
  \includegraphics[width=5.3in]{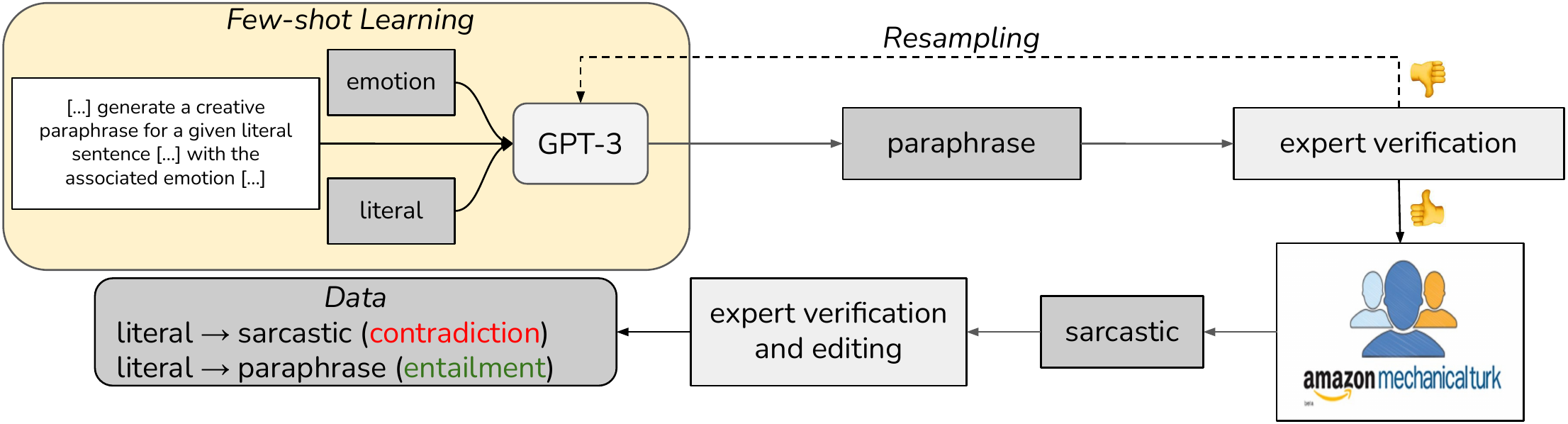}
  \caption{Model in the Loop for \flute: Sarcasm Data}
  \label{humanai} 
 \end{subfigure}
 \label{fig:superfigure_architecture}
\end{figure*}

\subsubsection{Sarcasm} \label{subsubsection:sarcasm}

When asked to generate literal equivalents of sarcastic sentences crowdworkers on Amazon Mechanical Turk (MTurk) usually perform trivial rephrasings at word/phrase level \cite{ghosh-etal-2020-interpreting}, which can lead to NLI datasets for sarcasm understanding where LLMs can achieve near-human performance (95\%) due to simple lexical cues, such as negation or antonyms \cite{chakrabarty-etal-2021-figurative}. Additionally, in many cases sarcasm data is collected from Twitter using hashtags, e.g., \#sarcasm, which can be noisy and not diverse. 

To address these issues we take a model-in-the-loop approach: given a literal sentence we first use GPT-3 with few-shot prompting to generate a literal paraphrase, and then use  crowdworkers to minimally edit this new literal sentence to form a sarcastic one (Figure \ref{humanai}).\footnote{Using GPT-3 to directly generate sarcastic sentences led to low quality output (no semantic consistency, social biases, stereotypes, or toxic content.} Then we pair the original literal sentence with generated literal paraphrase as entailment pair, and with the sarcastic one as a contradiction pair. Below we describe these two steps.

\paragraph{Entailment Pairs.}
Jointly modeling emotion and sarcasm had been shown beneficial for sarcasm detection \cite{chauhan-etal-2020-sentiment}. Thus, we select the literal sentences from the Empathetic Dialogue dataset \cite{rashkin-etal-2019-towards}. Each conversation in the dataset is grounded in a situation with emotion label provided.
We select literal sentences labeled with negative emotions such as \emph{angry}, \emph{afraid}, \emph{embarrassed}.\footnote{Although positive emotions can be used to generate sarcasm, we leave it for the future work.} Including the emotion in the GPT-3 prompts serves two purposes: (a) the generated paraphrases are complex and often more creative than the original literal input, and (b) it is easier for crowdworkers to transform paraphrases with emotional content into sarcastic counterparts with minimal edits. To generate the literal paraphrases, we provide the literal sentence and the associated emotion in the prompt and ask GPT-3 to paraphrase the input (top part of Figure \ref{humanai}, see prompt in Appendix \ref{sarcparaphrase}). Every paraphrase generated from GPT-3 is verified by 3 experts. If the quality of the generated paraphrase is deemed insufficient, it is resampled from the model. Any individual example undergoes at most three rounds of sampling, with 15\% of them judged as appropriate upon the first round. 

\paragraph{Contradiction pairs.} We recruit crowd workers on MTurk to convert the manually checked GPT-3-generated literal paraphrases into sarcastic sentences. The workers were provided with the paraphrases and instructed to make minimal edits (e.g., through negations, antonyms) to generate sarcastic sentences. We conducted a  qualification test, and recruited 29 distinct workers from the original set of 85 workers.\footnote{with native English proficiency, location in the U.S. and a hit success rate of 96\%.} 
We recruit two independent workers for every paraphrase input. The resulting sarcastic outputs are verified by three experts. 25\% of instances were deemed insufficient quality and edited by the experts.  Consider the sarcasm example in Table \ref{table:data}. The literal input ``My next door \dots'' is the premise. The first step generated the paraphrase hypothesis by adding the implicitly stated emotion ``annoyed'' and paraphrasing. Next, Turkers modified the paraphrase to its sarcastic counterpart - by replacing "annoying" (the emotion word) to it antonym ``pleasant''. 

The final \flute sarcasm benchmark consists of 2,678 sarcastic sentences contradicted by the 1,339 seed literal sentences, as well as the 1,339 paraphrased sentences (entailment pairs), as seen in Table \ref{table:data_stat}.

\subsubsection{Simile} \label{simile}

\paragraph{Entailment Pairs.} We start by extracting sentences containing similes from \cite{10.1162/tacl_a_00478,chakrabarty-etal-2021-figurative}. 
To generate an entailment pair, we perform two steps (see Fig. \ref{fig:architecture_allfig}): 1) given the sentence that contains a simile, create an auxiliary literal sentence by simply replacing the simile with the simile's property (e.g., replace \emph{like a steam sauna} with \emph{hot and moist}); 2) given the auxiliary literal sentence and the property, prompt GPT-3 to generate a \emph{literal paraphrase} consistent with the property (see prompt in Appendix \ref{prompt:simile_para}). Experts deemed 720 generated instances satisfactory. To illustrate the pipeline, see the example for simile in Table \ref{table:data}: ``In fact the assembly hall was now like a steam sauna''. The object \emph{steam sauna} is replaced with its property \emph{hot and moist}, and the resulting sentence is fed into GPT-3 to generate the paraphrase ``The assembly hall was now hot and moist, more so than usual’’.
\vspace{-1mm}
\paragraph{Contradiction pairs.} To generate the contradictions, we prompt GPT-3 to invert the meaning of the \emph{literal paraphrase} (see above) w.r.t the property (see prompt in Appendix \ref{prompt:simile_contra}). Out of 720 generated contradictions, experts deemed 642 satisfactory. We further selected 108 challenging $<$simile, entailment, contradiction$>$ instances from \newcite{https://doi.org/10.48550/arxiv.2204.12632}, defined by low RoBERTa \cite{zhuang-etal-2021-robustly} logit of the correct option (see Appendix \ref{subsubsection:simchallenge} for details), for a total of 750 entailment and 750 contradiction pairs (Table \ref{table:data_stat}).

\subsubsection{Metaphors} \label{subsubsection:metaphor}

\paragraph{Entailment Pairs.}  We follow a similar model-in-the-loop approach as the one for similes (Figure \ref{fig:architecture_allfig}). We manually select a total of 750 metaphors from the following datasets: \cite{chakrabarty-etal-2021-figurative,bigbench,stowe-etal-2022-impli}. Next, we prompt GPT-3 to generate paraphrases given the metaphoric sentences (see prompt in Appendix \ref{prompt:metaphor_para}). Although the original datasets contain the literal equivalents of the metaphoric sentence, they are not fully adequate for our purposes because: (a) not all metaphor examples have the literal counterpart, and  (b) often, the literal counterpart is a minimal modification (one-word edit) of the metaphor \cite{bigbench,chakrabarty-etal-2021-figurative}, which can lead to trivial examples for LLMs.

\paragraph{Contradiction Pairs.} 

To generate contradictions pairs, we start with the GPT-3 generated literal sentence that entails the metaphoric sentence. Consider the metaphor in the Table \ref{table:data}, ``He \emph{absorbed} the knowledge or beliefs of his tribe'' (taken from \cite{stowe-etal-2022-impli}). In the original dataset, the  non-entailment counterpart is ``He \emph{absorbed} the beverages of his tribe'', which is created by using a verb in a different sense (literal sense of ``absorb'') to fit the context of beverage drinking. On the contrary, since we are interested in generating instances that contradict the \emph{metaphor itself}, 
a  more appropriate  modification would be ``He \emph{utterly} \emph{decimated} his tribe’s most deeply held beliefs'' (more examples are in Table \ref{table:metaphor_ent_ex} in the Appendix). We follow the same method to generate contradiction examples (using GPT-3) as for similes (see prompt in Appendix \ref{prompt:metaphor_contra}).

Both paraphrases and contradictions are verified by three experts and edited when required. Our \flute benchmark contains 750 entailment and 750 contradictions pairs for metaphors (Table \ref{table:data_stat}). 

\subsubsection{Idioms} \label{subsubsection:idiom}
Observing the successful generations of paraphrases and contradictions by GPT-3 in the case of simile and metaphors, we jointly generate paraphrases along with contradictions using GPT-3 (See Figure \ref{fig:architecture_allfig} blue dotted lines). We provide the idiom and its meaning in the prompt (see prompt in Appendix \ref{idiomentailcontra}). Three experts manually verified all the generated sentences and edited a total of 23\% of total generations. We found that jointly generating paraphrases and contradictions greatly eased the data creation process and resulted in relatively high quality of generations. 
 
 \flute benchmark consists of 1000 entailment and 1000 contradiction pairs for idioms (Table \ref{table:data_stat}). 

\begin{table}[t]
\small
\centering
\begin{tabular}{|l|l|l|l|}
\hline
         & Entails    & Contradicts    & Total \\ \hline
Paraphrase & 1339 & - & 1339 \\ \hline         
+ Sarcasm  & - & 2678 & 2678  \\ \hline\hline
Simile   & 750  & 750  & 1500  \\ \hline
Metaphor & 750  & 750  & 1500  \\ \hline
Idiom & 1000  & 1000  & 2000  \\ \hline

\end{tabular}

\caption{\label{table:data_stat}Dataset statistics showing distribution of Figurative Language across \flute.}
\end{table} 

\subsection {\flute: Generating Textual Explanations} \label{section:NLE}
Our task prediction requires that the model not only correctly infer the label, but also explain \emph{why} a given premise entails or contradicts the hypothesis. 
Towards this goal, we generate textual explanations for every $<$premise, hypothesis$>$ pair.

For simile, metaphor, and sarcasm we provide the premise, hypothesis, and label (entailment or contradiction) and prompt GPT-3 to generate an explanation. We provide a natural language instruction followed by several examples. We generate entailment and contradiction explanations separately. 

For idioms, 
the idiom meaning in the seed dataset already makes up for a great explanation. Thus, we utilize the provided idiom meaning to jointly generate the explanation for the entailing premise, 
as well as for the contradicting premise 
using GPT-3. Hence, for idiom data in addition to premise, hypothesis, and labels, we also provide the idiom itself and its meaning in the prompt. 

Three experts manually verified all explanations to ensure their correctness and ability to explain the essence of the entailment or contradiction in reasonable detail rather then learning a simple template (see Table \ref{table:data}). 
In cases where explanations were not accurate, experts edited them to ensure they are coherent, logically consistent, and grammatical. For explanations pertaining to sarcasm+paraphrase, experts edited a total of 21\% of the generated explanations, while for simile, metaphor and idiom it was 27\%, 40\% and 10\% respectively, which further demonstrates the potential of using GPT-3 to significantly reduce the human effort that goes into collecting textual explanations datasets. See Appendix \ref{sec:appendix} for details on hyperparameters and prompts.

\section{Experimental Setup}

\subsection{Models} \label{section:models}
Prior works in explainability have trained two types of models. \textit{Pipeline} models map an input to a rationale (I $\rightarrow$ R), and then a rationale to an output (R $\rightarrow$ O).\footnote{Rationales are ``textual explanation'' in this work, sometime used interchangeably.} \textit{Joint Self Rationalizing} models  map an input to an output and rationale (I $\rightarrow$ OR). Recently \citet{wiegreffe-etal-2021-measuring} have exposed the short-comings of free-text pipelines and have empirically shown that joint model rationales are more indicative of labels. Following this, we fine-tune a joint self-rationalizing T5 model. Taking advantage of the text-to-text format of T5 \cite{raffel2020exploring} and the recent success of instruction-based models \cite{sanh2021multitask,wei2021finetuned}, we design the following instruction for a given literal premise (P) and a figurative hypothesis (H):

\textit{Does the sentence \textbf{{\textcolor{blue}{"P"}}} entail or contradict the sentence \textbf{{\textcolor{blue}{"H"}}}? Please answer between \textbf{{\textcolor{blue}{"Entails"}}} or \textbf{{\textcolor{blue}{"Contradicts"}}} and explain your decision in a sentence.}

The above instruction is fed to the encoder of T5. The decoder outputs the label followed by the rationale. We fine-tune T5 with the following setups: in the first one, we fine-tune on e-SNLI \cite{NEURIPS2018_4c7a167b}, and in the second, we fine-tune on \flute.

\textbf{T5$_{\text{e-SNLI}}$}: e-SNLI \cite{NEURIPS2018_4c7a167b} dataset comes with supervised ground-truth labels and rationales. We fine-tune the 3B version of T5 on e-SNLI for one epoch with a batch size of 1024, and an AdamW Optimizer with a learning rate of $1e-4$. We remove the \textit{Neutral} examples from e-SNLI because our test data does not have such a category. We take the longest explanation per example in e-SNLI since our data  has only one reference explanation. In case the explanations are more than one sentence we join them using `and' since our data contains single-sentence explanations. This leaves us with 366,603 training and 6,607 validation examples.

\textbf{T5$_{\flute}$}: We fine-tune the 3B version of T5 model for 10 epochs with a batch size of 1024, and an AdamW Optimizer with a learning rate of $1e-4$ in a multitask fashion with data from all the four types of figurative languages combined. Our training data consists of 7,035 samples which is 50X smaller than e-SNLI. For validation we use 500 examples which is used for selecting best checkpoint based on loss.

\subsection{Evaluation Setup} \label{section:eval}

To evaluate the above models, we built a test set by randomly selecting 750 instances (i.e., $<$premise, hypothesis$>$ pairs with associated explanations) from the sarcasm dataset, and 250 examples each from simile, metaphor and idiom datasets, for a total of 1,500 instances.

Below we describe several automatic metrics and human evaluations we consider to assess the models' ability to understand figurative language.

\paragraph{Automatic Metrics} To judge the quality of the explanations we compute the average between BERTScore \cite{Zhang-etal:2020:bertscore} \footnote{We use the deberta-mnli version that has shown to have highest correlation with human judges.} and BLEURT \cite{sellam-etal-2020-bleurt}, which we refer to as \emph{explanation score} (between 0 and 100). Instead of reporting only label accuracy, we report label accuracy at three thresholds of explanation score (0, 50, and 60). Accuracy@0 is equivalent to simply computing label accuracy, while Accuracy@50 counts as correct only the correctly predicted labels that
achieve an explanation score greater than 50.
\paragraph{Rationale Quality} Human simulatability \cite{doshi2017towards} has a rich history in machine learning interpretability research as a reliable measure of rationale quality from the lens of utility to an end-user. Simulatability measures the additional predictive ability a rationale R provides over the input I for a given label O, computed as the difference between task performance when a rationale is given as input vs. when it is not (IR $\rightarrow$ O minus I $\rightarrow$ O).In prior work on Explanability using Natural Language \citet{wiegreffe-etal-2021-measuring} pointed out that model predictions are often unable to be simulated because they degenerate under high values of noise. Following \citet{wiegreffe-etal-2021-measuring} we thus use a variant of this metric that relies on predicting the gold labels as our measure of rationale quality: (IR $\rightarrow$ \^{O} minus I $\rightarrow$ \^{O}).

To compute rationale quality, we first train IR $\rightarrow$ O and I $\rightarrow$ O for both \flute and e-SNLI data. We then compute the test accuracy for \flute for both the IR $\rightarrow$ O and I $\rightarrow$ O models trained on e-SNLI and \flute using predicted rationales from the respective I $\rightarrow$ OR models. We also compute rationale quality with gold rationales (R*).

\begin{table*}[t]
\centering
\renewcommand{\arraystretch}{1.2}
\begin{tabular}{lcccccccccccc}
    & \multicolumn{6}{l}{T5$_{\text{e-SNLI}}$}                                                                                                                                                                    & \multicolumn{5}{l}{T5$_{\flute}$}                                                                                                                                            &   \\\hline
    & \begin{tabular}[c]{@{}l@{}}Acc\\@0\end{tabular} & \begin{tabular}[c]{@{}l@{}}Acc\\ @50\end{tabular} & \begin{tabular}[c]{@{}l@{}}Acc\\ @60\end{tabular} & H$_{\text{score}}$ & Yes\% & \multicolumn{1}{l|}{No\%} & \begin{tabular}[c]{@{}l@{}}Acc\\ @0\end{tabular} & \begin{tabular}[c]{@{}l@{}}Acc\\ @50\end{tabular} & \begin{tabular}[c]{@{}l@{}}Acc\\ @60\end{tabular} & H$_{\text{score}}$ & Yes\% & No\% \\\hline
\small{Sarcasm} & 60.6                                              & 15.7                                               & 2.4                                                &  34.2    & 14.7  &  \multicolumn{1}{l|}{52.0}   & \textbf{91.6}                                              & \textbf{86.2}                                               & \textbf{56.2}                                               &   \textbf{85.3}   & \textbf{75.3}   & \textbf{8.7}  \\
\small{Simile} & 61.2                                              & 22.8                                               & 3.6                                                &  43.6    &  22.0 & \multicolumn{1}{l|}{40.7}  & \textbf{62.8}                                              & \textbf{57.2}                                               & \textbf{30.4}                                               & \textbf{84.9} & \textbf{74.7} &  \textbf{8.0} \\
\small{Metaphor} & \textbf{81.8}                                              & 31.8                                               & 11.6                                               &  55.3    &  36.0 & \multicolumn{1}{l|}{28.0}  & 73.3                                              & \textbf{55.6}                                              & \textbf{23.7}                                               &  \textbf{80.2}    & \textbf{64.0}  & \textbf{6.0}  \\
\small{Idiom} & \textbf{84.8}                                              & 46.4                                               & 7.6                                                &  60.9    &  37.3 & \multicolumn{1}{l|}{24.7}  & 79.2                                              & \textbf{77.2}                                               & \textbf{66.8}                                              &  \textbf{83.1}    & \textbf{69.3}  &  \textbf{8.7}
\\\hline
\end{tabular}
\caption{\label{results} Accuracy scores across four figurative language types by varying thresholds of explanation score, along with human evaluation scores H$_{\text{score}}$, Yes\% (higher is better), and No\% (lower is better) for explanations generated by T5 fine-tuned on e-SNLI (T5$_{\text{e-SNLI}}$) and T5 fine-tuned on \flute ({T5$_{\flute}$}). $p<0.001$ via Wilcoxon signed-rank test for all bolded results.}
\end{table*}

\paragraph{Human Evaluation.} \label{subsection:humaneval}
Finally, we measure the quality of the generated textual explanations from T5$_{\text{e-SNLI}}$ and T5$_{\flute}$ models via the MTurk platform. We recruit 79 crowd workers with at least 98\% HIT approval rate. We compute human judgement scores (H$_{\text{score}}$), identical to the e-ViL score in  \newcite{Kayser_2021_ICCV}. For each NLI instance (a total of 200 random samples, 50 per figurative language type), we present two textual explanations generated by the two models (T5$_{\text{e-SNLI}}$ and T5$_{\flute}$) and ask three workers the following question: \textit{Given the two sentences, does the explanation justify the answer above?} We provide four options: \emph{Yes} ($1$), \emph{Weak Yes} ($\frac{2}{3}$), \emph{Weak No} ($\frac{1}{3}$), and \emph{No} ($0$). For each explanation, we average the scores by the three annotators and report the sample average in Table \ref{results} as H$_{\text{score}}$. If the answer is anything other than \emph{Yes}, we ask to categorize the shortcomings of the explanation:  \emph{Insufficient Justification}, \emph{Too Trivial}, \emph{To Verbose}, \emph{Untrue to Input}, \emph{Violates Common Sense} \cite{majumder2021rationale}.

Three workers were recruited for each instance and the IAA using Krippendorff’s $\alpha$ \cite{krippendorff2011computing} between the workers is $0.45$, indicating moderate agreement. See Appendix \ref{appendix:humaneval_details} for details on the H$_{\text{score}}$ computation and Figure \ref{fig:mturk} for a screenshot of the MTurk task interface. 

\section{Results and Discussion} 
\label{section:result}

Table \ref{results} shows accuracy at varying \emph{explanation score} thresholds. A threshold of 0\% does not account for the quality of the textual explanation and is equivalent to simply reporting label accuracy. With an increase in threshold to greater than 50\% we see accuracy scores dropping almost by half for T5$_{\text{e-SNLI}}$, showing most explanations generated from the model trained on e-SNLI are of poor quality. By increasing the threshold to greater than 60\%, the accuracy scores further decrease, demonstrating that models like T5 fine-tuned on e-SNLI still struggle generating correct explanations even when the label predictions are correct. On the contrary, Table \ref{results} shows that the accuracy scores for T5$_{\flute}$ are significantly higher for each type of figurative language, indicating higher quality of explanations achieved by fine-tuning the model on our dataset. 

\begin{table}[h]
\centering
\small
\begin{tabular}{|l|l|l|l|}
\hline
      & Ac (IR $\rightarrow$ O)  & Ac (I $\rightarrow$ O) & RQ. \\ \hline
e-SNLI & 68.4 & 74.5   & -6.1    \\ \hline
\flute &  89.3    &  90.5  & \textbf{-1.2}    \\ \hline\hline
\flute(R*) &  95.6    &  90.5  & \textbf{5.1}    \\ \hline
\end{tabular}
\caption{\label{sim}Rationale Quality ($p<0.001$ via Wilcoxon signed-rank test) (higher is better) on FLUTE test set using accuracy of IR $\rightarrow$ O  and I $\rightarrow$ O models trained on e-SNLI and \flute respectively. We use the predicted rationale R obtained from respective I $\rightarrow$ OR models T5$_{\text{e-SNLI}}$ and T5$_{\flute}$ as well as gold rationale (R*). Rationale Quality and accuracy are abbreviated as RQ. and Ac.}
\end{table}

In terms of the Rational Quality (Table \ref{sim}), using predicted rationales from I $\rightarrow$ OR we observe that model fine-tuned on e-SNLI generates explanations that are less useful for gold label prediction compared to a model trained on \flute. Simulatability scores obtained using predicted rationales are negative compared to gold rationales since rationales generated by the model are still imperfect and may introduce noise that negatively affects the prediction. The positive score of gold rationales indicates that explanations provided in the \flute dataset are indeed useful for the model to make predictions.

\begin{table}[h]
\centering
\small
\renewcommand{\arraystretch}{1.2}
\begin{tabular}{|p{1.23cm}|llll|}
\hline
Premise    & \multicolumn{4}{l|}{\begin{tabular}[c]{@{}l@{}}The camp life really had energized him.\end{tabular}}                                                            \\ \hline
Hypothesis & \multicolumn{4}{l|}{\begin{tabular}[c]{@{}l@{}}Life in the camp wilted him.\end{tabular}}                                                                  \\ \hline
Label & \multicolumn{4}{l|}{\begin{tabular}[c]{@{}l@{}}Contradiction.\end{tabular}}                                                                   \\ \hline
Gold\_Exp       & \multicolumn{4}{l|}{\begin{tabular}[c]{@{}l@{}}To wilt means to lose vitality or strength,\\ while to energize means to give strength\\ or energy to.\end{tabular}}    \\ \hline
T5$_{\text{e-SNLI}}$    & \multicolumn{4}{l|}{\begin{tabular}[c]{@{}l@{}}If the camp life energized him, then\\ he is not wilted.\end{tabular}}                                                            \\ \hline
T5$_{\flute}$   & \multicolumn{4}{l|}{\begin{tabular}[c]{@{}l@{}}To wilt something means to lose its energy or\\ vitality, while to have energized something \\mean to feel a sense of excitement or life.\end{tabular}} \\ \hline\hline
Premise    & \multicolumn{4}{l|}{\begin{tabular}[c]{@{}l@{}}No one came to my wedding.\end{tabular}}                                                            \\ \hline
Hypothesis & \multicolumn{4}{l|}{\begin{tabular}[c]{@{}l@{}}No big deal that there were no guests at my\\ wedding reception considering it's the only \\one I'll ever have.\end{tabular}}                                                                   \\ \hline
Label & \multicolumn{4}{l|}{\begin{tabular}[c]{@{}l@{}}Contradiction.\end{tabular}}                                                                   \\ \hline
Gold\_Exp       & \multicolumn{4}{l|}{\begin{tabular}[c]{@{}l@{}}A wedding reception is supposed to be a \\joyous event with many guests, so not \\having any guests at it would make it a \\very uneventful and pointless one.\end{tabular}}    \\ \hline
T5$_{\text{e-SNLI}}$    & \multicolumn{4}{l|}{\begin{tabular}[c]{@{}l@{}} {\color{red}No one is same as No big deal}.\end{tabular}}                                                            \\ \hline
T5$_{\flute}$   & \multicolumn{4}{l|}{\begin{tabular}[c]{@{}l@{}}A wedding reception is a celebration of the\\ couple's union and is usually attended by \\family and friends so not having guests \\there would be seen as snub.\end{tabular}} \\ \hline
\end{tabular}
\caption{\label{modelexpl}Examples of T5$_{\text{e-SNLI}}$ and T5$_{\flute}$  model generated explanations vs. gold explanations for NLI involving metaphor (top) and sarcasm (bottom). More examples in Table \ref{examples} in Appendix.}
\end{table}
\vspace{-1mm}

\begin{figure}[h]
    \centering
    \includegraphics[width=\columnwidth]{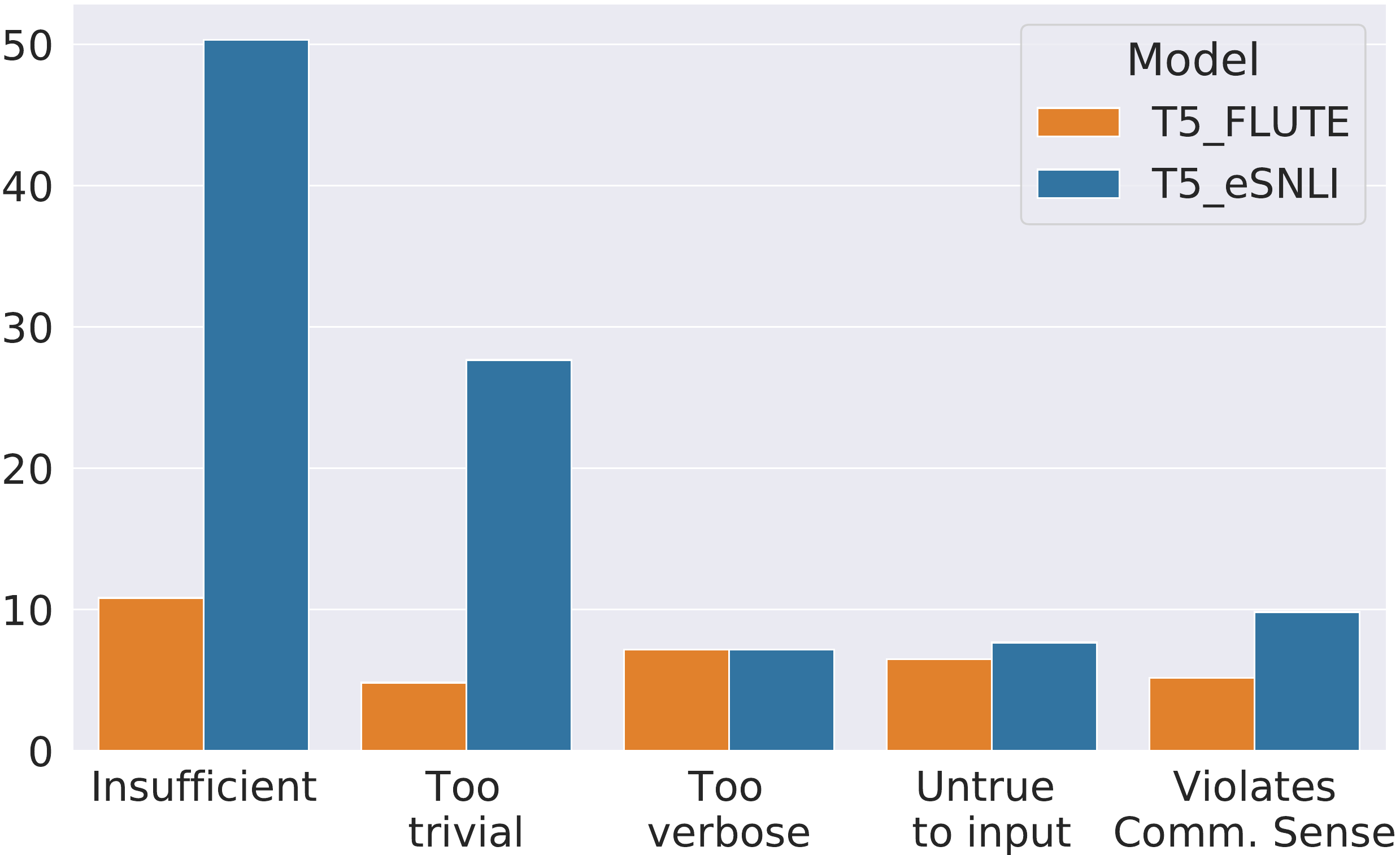}
    \caption{Bar plot of the number of crowd worker-identified shortcomings of explanations generated by T5$_{\text{e-SNLI}}$ and T5$_{\flute}$ by shortcoming and model type as percent of the sample (lower means fewer shortcomings). An extended plot by figurative language type is available in Appendix, Figure \ref{fig:shortcoms_by_figtype}.}
    \label{fig:shortcomings_bar}
\end{figure}

Table \ref{results} also presents the H$_{\text{score}}$ scores for explanations collected from human evaluation (see Appendix \ref{appendix:humaneval_details} and prior work \cite{Kayser_2021_ICCV, majumder2021rationale} for details on computation). We notice the scores for T5$_{\flute}$ are 51.1, 41.3, 24.9, 22.2 points better than for T5$_{\text{e-SNLI}}$ for Sarcasm, Simile, Metaphor, Idiom (in that order). Crowd workers answered with solid `Yes' (that the explanation justifies the label) in 43.4\% more cases on average when presented explanations from T5$_{\flute}$ compared to T5$_{\text{e-SNLI}}$. Likewise, they answered `No' in 28.5\% less cases on average when presented explanations from T5$_{\flute}$ compared to T5$_{\text{e-SNLI}}$. These results demonstrate a notably higher quality of explanations generated by the model fine-tuned on \flute compared to the model fine-tuned on e-SNLI, despite the significantly larger size of the latter dataset. 

Figure \ref{fig:shortcomings_bar} shows that for cases where crowd workers did not choose an absolute 'Yes', the most common error category to be found is \textit{Insufficient Justification}. The percentage of explanations for which shortcomings were identified is higher or the same for across all categories for T5$_{\text{e-SNLI}}$. The second example (``No one is same \dots'') in Table \ref{modelexpl} shows an insufficient explanation generated by T5$_{\text{e-SNLI}}$. T5$_{\text{e-SNLI}}$ generated explanations were also more frequently marked as \textit{Too Trivial}. Often, they do not explain the reasoning but rather follow a standard template \textit{if A then not B} or \textit{if A then B}, such as the first example in Table \ref{modelexpl} for T5$_{\text{e-SNLI}}$. We share more such examples of erroneous explanations in the Appendix, Table \ref{table:shortcomings_ex}.

\section{Related Work} \label{section:re}

In recent years, evaluating how well RTE models can capture specific linguistic phenomena such as figurative language has attracted many NLP researchers. In earlier work, \cite{agerri-2008-metaphor} analyze metaphor examples in the RTE-1 \cite{dagan2006pascal} and RTE-2 \cite{BarHaim2006TheSP} datasets,  whereas \cite{poliak-etal-2018-collecting-diverse}'s diverse collection of RTE examples contains one type of figurative language - puns. Our research is closer to \cite{chakrabarty-etal-2021-figurative}, however, \flute has better diversity and also contains explanations for each NLI instance, as well as multiple expert checks to ensure higher quality (see Section \ref{sec:data}).  A portion of \flute's metaphors are based on the Big-bench corpora \cite{bigbench} as well as the IMPLI dataset \cite{stowe-etal-2022-impli}, which is inspired by \cite{zhou-etal-2021-pie}'s prior work on the paired idiomatic and literal dataset. However, there are several distinctions between these datasets and \flute. First, not all the metaphors in Big-bench have literal paraphrase and contradictions. Second, in case of both the Big-bench and IMPLI dataset, the non-entailment examples are created via minimal edits to the original metaphor, often resulting in neutral examples or contradiction to the non-metaphoric part of the sentence. In \flute, we ensure that the non-entailment examples are in contradiction to the metaphor (Table \ref{table:metaphor_ent_ex}). 

One of the motives for having NLEs with <literal, figurative> sentence pairs like \flute does is to evaluate model's ability to explain their decisions. Recent datasets such as  CoS-E \cite{rajani-etal-2019-explain}, Movie Reviews \cite{zaidan-eisner-2008-modeling}, and e-SNLI \cite{NEURIPS2018_4c7a167b} have been released in a similar vein. Recent work has also leveraged large language models to explain humor in image captions \cite{hessel2022androids} or sarcasm in dialouges \cite{kumar-etal-2022-become}. The e-SNLI dataset (i.e., NLE of the entailment relations in the SNLI dataset) has been used in related work \cite{narang2020wt5,yordanov2021few,majumder2021rationale,feng-etal-2022-neuro} for explanation generation. In contrast to e-SNLI, which was created via crowdsourcing, we rely on a model-in-the-loop framework for \flute influenced by \cite{wiegreffe2021reframing}. We have utilized the e-SNLI dataset for explanation generation and observed a T5 model trained on \flute performs notably better.

\section{Conclusion}
We release \flute, a  dataset for figurative language understanding spanning across Sarcasm, Similes, Metaphors, and Idioms collected via a model-in-the-loop framework. To encourage genuine understanding of figurative language, our data also contains free-form textual explanations. Upon conducting baseline experiments with state-of-the-art benchmark models (i.e., models trained on the the e-SNLI dataset), we notice those models perform poorly. In contrast, performance of the T5 model fine-tuned on \flute shows that our dataset can bring us a step closer to developing models that understand figurative language through textual explanations. We hope our research on explanation generation for figurative language will be a fruitful future direction, and our dataset will be a challenging testbed for experimentation.

\section*{Limitations}
While we focused on four types of figurative language and generated a diverse dataset, we believe it is just a first step towards capturing figurative NLI instances and their explanations, since figurative language is able to draw on a wide variety of cultural knowledge and contexts. Although the sarcasm portion captures the most common type of incongruity between sarcastic context and sentiment, sarcasm can manifest in many different forms - situational, underplayed, or dramatic, for which examples and explanations will differ. Finally this study doesn't explicitly focus on faithfulness of model generated Natural Language Explanations, however we hope to evaluate the faithfulness of these using methods described in contemporaneous literature on faithfulness of NLE's \cite{sia2022logical,chan2022frame}
\section*{Ethics Statement}

We use a model-in-the-loop framework for content generation. Although we use language models trained on data collected from the Web, which have been shown to have issues with gender bias and abusive language, we have verified carefully that our FLUTE data does not  contain any toxic text and it underwent manual inspection by the authors and experts.We pay Amazon Mechanical Turkers at the rate of 15 \$/hr which is compliant with the minimum hourly wage in United States.

\section*{Acknowledgements}
We would like to thank the anonymous reviewers for their helpful comments. Finally we are also grateful to Esin Durmus and Faisal Ladakh for initial brainstorming on model in the loop strategies for dataset creation. Tuhin is funded by Columbia Center of Artifical Intelligence \& Technology (CAIT) and the Amazon Science Ph.D. Fellowship). 

\bibliographystyle{acl_natbib}
\bibliography{anthology,custom}
\clearpage
\appendix

\section{Appendix}
\label{sec:appendix}
In this section we report the details of the experiments (e.g., hyperparameters used for GPT-3 based generations, examples, as well as the prompts for the figurative language.

\subsection{Sarcasm dataset} \label{subsection:appsarc}

\subsubsection{Hyperparameters for the sarcasm dataset}

We use GPT-3-Davinci-001 model for auxillary paraphrase generations from which crowd workers create sarcasm.
To generate paraphrase, we use the following hyperparameters: \texttt{temperature=1,
max tokens=100,
top p=0.9,
best of=1,
frequency penalty=0.5,
presence penalty=0.5.}

To generate explanations, we use the following hyperparameters: \texttt{
temperature=1.0,
max tokens=100,
top p=0.9,
frequency penalty=0.5,
presence penalty=0.5,
stop=["."].}

\subsubsection{Prompts for generation of Paraphrase from which Sarcasm is created}
\label{sarcparaphrase}

\textit{You will be presented with examples of some literal input sentences and their creative paraphrases. For each example, we also provide the associated emotion. Your task is then to generate a creative paraphrase for a given literal sentence where the creative paraphrase should reflect the associated emotion without changing its meaning. Make sure to use some sort of humor and commonsense about everyday events and concepts}

\begin{itemize}
    \item[] 1) \textbf{Literal}: A lot of people have got engaged recently.
    \item[] \textbf{Emotion}: surprised
    \item[] \textbf{Creative Paraphrase}: The way all the couples are pairing off lately and naming the big day, I think Cupid's really busy.
    \\
    \item[] 2) \textbf{Literal}: We have enough candles mom
    \item[] \textbf{Emotion} annoyed
    \item[] \textbf{Creative Paraphrase}: I think the Catholic church is going to have to canonize a whole new generation of saints to justify our candle use mom\\
\dots

\end{itemize} 
'

\subsubsection{Prompts for generation of Explanation for Paraphrase from which Sarcasm is created (Entailment)}
\label{sarcaexplaent}
\textit{You will be presented with examples of two sentences typically a premise along with an entailing paraphrase of the premise called the hypothesis. Your task is to generate natural language explanations to justify the Entailment between the premise and the hypothesis.} 

\begin{itemize}
\item[] 1) \textbf{Premise}: Awful seeing a naked man run through my neighborhood.
\item[]\textbf{Hypothesis}: The sight of a man running through my neighborhood sans clothes was pretty disgusting.
\item[]\textbf{Explanation}: It is socially unacceptable to not wear clothes and step out of one's house so seeing a man who is running naked in the neighborhood is pretty shameful and disgusting.
\\
\item[] 2) \textbf{Premise}: My mother didn't cook her chicken all the way through at dinner the other night.
\item[] \textbf{Hypothesis}: The fact that my mother didn't cook her chicken all the way through at dinner makes me feel like I'm going to vomit.
\item[] \textbf{Explanation}: Eating undercooked chicken can cause food poisoning and so finding out that the chicken at dinner wasn't cooked all the way through often makes people throw up
\dots

\end{itemize}

\subsubsection{Prompts for generation of Explanation for Sarcasm (Contradiction)}
\label{sarcaexpla}
\textit{You will be presented with examples of some literal and sarcastic sentences. Your task is then to write explanations to justify why it is sarcastic w.r.t the literal} 

\begin{itemize}
\item[] 1) \textbf{Literal}: When I moved into my apartment it was full of bugs
\item[]\textbf{Sarcasm}: I absolutely loved when I moved into my apartment and found it crawling with bugs.
\item[]\textbf{Explanation}: Bugs are usually disgusting and most people are terrified of them therefore it is unlikely to love seeing someone's apartment infested by them.
\\
\item[] 2) \textbf{Literal}: I've been hearing some strange noises around the house at night.
\item[] \textbf{Sarcasm}: I am completely comforted by the weird noises I keep hearing around the house at night.
\item[] \textbf{Explanation}: Hearing weird noises around the house at night could invoke a potential danger such as a robbery or someone breaking in with malicious intent which makes someone scared rather than comforted.
\dots

\end{itemize}

\subsection{Simile dataset} \label{subsection:appsimile}

\subsubsection{Hyperparameters for the simile dataset}

We use GPT-3-Davinci-002 model for simile data generations.
To generate paraphrase, we use the following hyperparameters: \texttt{temperature=1,
max tokens=256,
top p=0.5,
best of=1,
frequency penalty=0.5,
presence penalty=0.1.}

To generate contradictions, we use the following hyperparameters: \texttt{temperature=0,
max tokens=100,
top p=1,
frequency penalty=0,
presence penalty=0.}

To generate explanations, we use the following hyperparameters: \texttt{
temperature=0.7,
max tokens=86,
top p=1,
frequency penalty=0,
presence penalty=0,
stop=["."].}

\begin{figure}[h]
    \centering
    \includegraphics[scale=0.8]{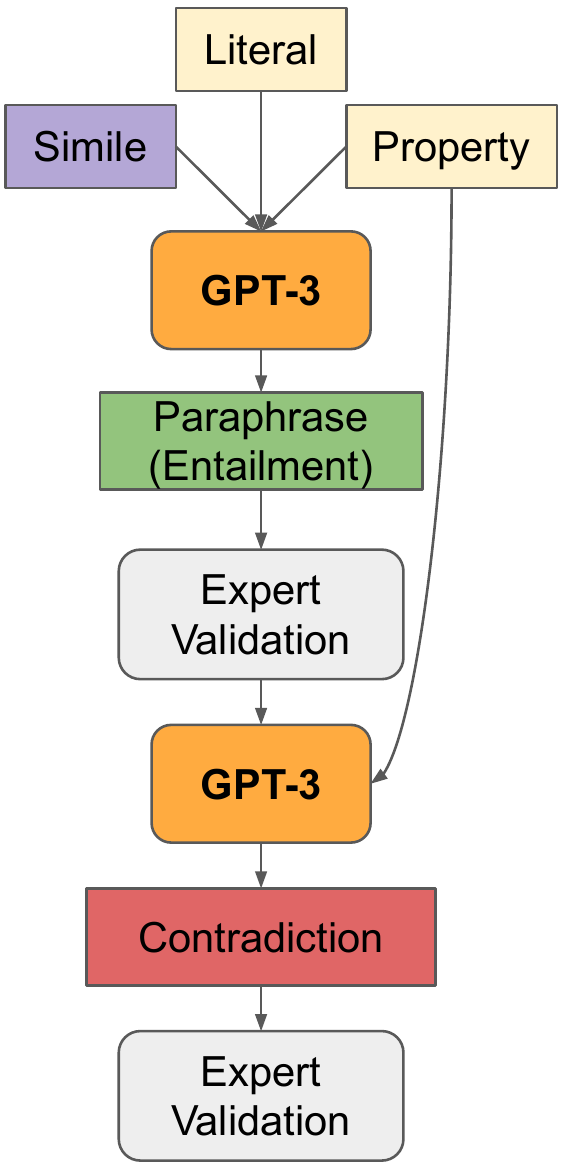}
    \caption{\label{simile_chart}Model-in-the-loop to generate a Simile NLI dataset.}
    \label{fig:simile_chart}
\end{figure}

\subsubsection{Challenging instances for the simile dataset} \label{subsubsection:simchallenge}
To select challenging instances from FigQA dataset \citet{https://doi.org/10.48550/arxiv.2204.12632}, we use RoBERTa fine-tuned on SNLI, MNLI, FEVER-NLI, and ANLI (R1, R2, R3). We choose <simile, literal, contradiction> instances by taking the average between RoBERTa logit for entailment given <simile, literal> as input and the logit for contradiction given the <simile, contradiction> input. We then choose instances with the lowest such score. In this way, we are able to select instances for which RoBERTa is essentially confusing entailment and contradiction. From our observation, these are usually ironic similes, e.g. for the simile \emph{I'm sharp as a pillow} and the literal sentence \emph{I'm not sharp}, RoBERTa would have a low logit for entailment, while for the sentence \emph{I'm sharp} it would have a low logit for contradiction.

\subsubsection{Prompts for Simile Paraphrase Generation}
\label{prompt:simile_para}
\textit{You will be presented with examples of some literal input sentences and their creative paraphrases. You will also be presented words that need to be preserved. Your task is to generate a creative paraphrase for a given literal sentence consistent in meaning. DO NOT CHANGE words after "Preserve:" keyword.}
\begin{itemize}
    \item[] 1. \textbf{Sentence:} overwhelmingly , it began to draw him in. \\
    \textbf{Preserve:} overwhelmingly\\
    \textbf{Creative Paraphrase:} He was overwhelmingly obsessed with it.
    \\
    .
    .
    .
\end{itemize}

\subsubsection{Prompts for Simile Contradiction Generation}
\label{prompt:simile_contra}
\textit{You will be presented with a Sentence and a Property. Your goal is to invert the meaning of the Sentence with respect to the Property via a minimal edit.}
\begin{itemize}
    \item[] 1. \textbf{Sentence:} The place looked impenetrable and inescapable. \\
    \textbf{Property:}  impenetrable and inescapable\\
    \textbf{Inversion:} This place looked easy to walk into and exit from.
    \\
    .
    .
    .
\end{itemize}

\subsubsection{Prompts for Simile Contradiction Explanation Generation}
\label{prompt:simile_expl}
\textit{You will be provided with a Simile and a contradictory sentence after the word "Contradiction". Your task is to explain why the contradictory sentence contradicts the Simile.}
\begin{itemize}
    \item[] 1. \textbf{Simile:}  like a psychic whirlpool , it began to draw him in. \\
    \textbf{Contradiction:}  Mildly, it began to draw him in\\
    \textbf{Explanation:} A whirlpool is a strong current, so a psychic whirlpool drawing in indicates that it was drawing him in intensely, rather than mildly.
    \\
    .
    .
    .
\end{itemize}

\subsection{Prompts for Simile Entailment Explanation Generation}
\label{prompt:simile_expl_ent}
\textit{You will be presented with a sentence containing a simile (Simile Sentence) and an entailing sentence (Entail Sentence). Please provide an explanation for why Simile Sentence is implied by the Entail Sentence.}
\begin{itemize}
    \item[] 1) \textbf{Simile Sentence:}   The place looked like a fortress \\
    \textbf{Entail Sentence:} The place looked impenetrable and inescapable\\
    \textbf{Explanation:} A fortress is a military stronghold, hence it would be very hard to walk into, or in other words impenetrable.
    \\
    .
    .
    .
\end{itemize}

\subsection{Metaphor dataset} \label{subsection:appmet}

\subsubsection{Hyperparameters for the metaphor dataset}

We use GPT-3-Davinci-002 model for metaphor data generations.
To generate paraphrase, we use the following hyperparameters: \texttt{temperature=1,
max tokens=100,
top p=0.8,
best of=1,
frequency penalty=0.5,
presence penalty=0.1.}

To generate contradictions, we use the following hyperparameters: \texttt{temperature=0,
max tokens=100,
top p=0.8,
frequency penalty=0.5,
presence penalty=0.1.}

To generate explanations, we use the following hyperparameters: \texttt{
temperature=0.8,
max tokens=100,
top p=0.9,
frequency penalty=0,
presence penalty=0,
stop=["."].}

\subsubsection{Prompts for Metaphor Paraphrase Generation}
\label{prompt:metaphor_para}
\textit{You will be presented with examples of some metaphor input sentences and their creative paraphrases. Your task is to generate a creative paraphrase for a given literal sentence consistent in meaning.}
\begin{itemize}
    \item[] 1. \textbf{Sentence:} A golden sun shines high in the sky. \\
    \textbf{Creative Paraphrase:} a very bright sun shines high in the sky.
    \\
    .
    .
    .
\end{itemize}

\subsubsection{Prompts for Metaphor Contradiction Generation}
\label{prompt:metaphor_contra}
\textit{You will be presented with examples of some literal input sentences and their contradictions. Your task is then to generate contradiction of the new sentences via a minimal edit.}
\begin{itemize}
    \item[] 1. \textbf{Sentence:} The company released him after many years of service. \\
    \textbf{Contradiction:} The company hired him after many years of service.
    \\
    .
    .
    .
\end{itemize}

\subsubsection{Prompts for Metaphor Entailment Explanation Generation}
\label{prompt:metaphor_entail_expl}
\textit{You will be presented with examples of sentences containing a metaphor along with an entailing paraphrase of the sentence. Your task is to generate natural language explanations to justify the Entailment with the input.}
\begin{itemize}
    \item[] 1. \textbf{Metaphor:}  Krishna is an early bird. \\
    \textbf{Contradiction:}  Krishna wakes up early everyday.\\
    \textbf{Explanation:} Early bird means the person who wakes up early in the morning.
    \\
    .
    .
    .
\end{itemize}

\subsubsection{Prompts for Metaphor Contradiction Explanation Generation}
\label{prompt:metaphor_contra_expl}
\textit{You will be provided with a sentence containing a Metaphor and a contradictory sentence after the word ``Contradiction". Your task is to explain why the contradictory sentence contradicts the Metaphor.}
\begin{itemize}
    \item[] 1. \textbf{Metaphor:}  Joseph has the heart of a lion. \\
    \textbf{Contradiction:}  Joseph has the calm demeanor of a lamb.\\
    \textbf{Explanation:} A lamb is typically seen as a gentle and timid creature while a lion is seen as a brave and fierce creature.
    \\
    .
    .
    .
\end{itemize}

\begin{table}[htbp]
\centering
\small
\begin{tabular}{ |p{2cm}|p{5cm}| }
\hline
Metaphor & A weather vane \emph{crowns} the building. \\
Original Entailment & The king crowned the prince. \\
GPT-3 Contradiction & A weather vane mars the building. \\
\hline
Metaphor & A \emph{golden} sun shines high in the sky.\\
Original Entailment & A very expensive sun shines high in the sky	A sunset is setting in the west. \\
GPT-3 Contradiction & A sunset is setting in the west. \\
\hline
Metaphor & Fear had changed him to a \emph{shaken jelly}. \\
Original Entailment & He was afraid of shaking jelly. \\
GPT-3 Contradiction & He had conquered his fear and now he is a strong and capable person. \\
\hline
\end{tabular} 
\caption{Example of Metaphors and their contradictions (from prior work and generated for this paper.). Note, examples from prior work replace the metaphor sentence with adding words that fits into their context whereas in this work we generate examples that \emph{contradicts} the metaphor.}
\label{table:metaphor_ent_ex}
\end{table}

\subsection{Idiom dataset} \label{subsection:appidiom}

\subsubsection{Hyperparameters for the idiom dataset}

We use GPT-3-Davinci-002 model for idiom data generations.
To jointly generate paraphrase and contradiction, we use the following hyperparameters: \texttt{temperature=1,
max tokens=200,
top p=0.9,
best of=1,
frequency penalty=0.5,
presence penalty=0.5, 
stop=[".."].}

To jointly generate explanations for entailment and contradiction, we use the following hyperparameters: \texttt{
temperature=0.7,
max tokens=256,
top p=0.9,
frequency penalty=0,
presence penalty=0,
stop=[".."].}

\subsection{Prompts for joint paraphrase and contradiction generation for idioms}
\label{idiomentailcontra}
\textit{You will be presented with examples of some input sentences containing an idiom. You will be provided with the meaning of the idiom. Your task is to first generate a paraphrase that complies with the meaning of the idiom and then generate a negation of the paraphrase that contradicts the meaning of the idiom. Please look at the span within bold tags when performing paraphrase and negation.}
\begin{itemize}
    \item[] 1) \textbf{Sentence}: He looked great, and he was smiling <b>to beat the band</b>.
    \item[] \textbf{Idiom}: to beat the band
\item[] \textbf{Meaning}: To a huge or the greatest possible extent or degree.
\item[] \textbf{Paraphrase}: He looked awesome and was smiling <b>hilariously in an uncontrollable manner</b>
\item[] \textbf{Negation}: He looked awesome and was smiling <b> in a very coy and restrained manner</b>.
\\
\dots
    
\end{itemize}
\subsubsection{Prompts for joint generation of Explanation for idioms}
\label{idiomexpla}
\textit{You will be presented with examples of sentences containing an idiom along with an entailing and contradictory paraphrase of the sentence. Your task to generate natural language explanations to justify the Entailment or Contradiction with the input.}

\begin{itemize}
    \item[] 1) \textbf{Sentence}: Not to share the bank with the table, or to take some minor part of it, but to go the whole hog.
    \item[] \textbf{Idiom}: go the whole hog
    \item[] \textbf{Meaning}: To do something as thoroughly as possible or without restraint.
    \item[] \textbf{Entailment}: Not to share the bank with the table, or to take some minor part of it, but to take it all for themselves without any restraint.
    \item[] \textbf{Contradiction}: Not to share the bank with the table, or to take some minor part of it, but to show some restraint and not go overboard.
    \item[] \textbf{Entail\_Explanation}: Usually to go the whole hog refers to do something as thoroughly as possible , taking it all for oneself or without any restraint.
    \item[] \textbf{Contra\_Explanation}: Usually to go the whole hog refers to do something as thoroughly as possible, without any sort of restraint and is often characterized by being extreme or overboard..
    \\
    \item[] 2) \textbf{Sentence}: I told her almost everything, including my reticence about seeing this work of literature go through the mill which was vanity publishing.
Idiom: go through the mill
    \item[] \textbf{Idiom}: go through the mill
    \item[] \textbf{Meaning}: To be abused or treated very harshly; to suffer intense anguish, stress, or grief.
    \item[] \textbf{Entailment}: I told her almost everything, including my reticence about seeing this work of literature be abused and treated in an extremely harsh manner which was vanity publishing.
    \item[] \textbf{Contradiction}: I told her almost everything, including my reticence about seeing this work of literature be celebrated and treated in an extremely proper manner which was vanity publishing.
    \item[] \textbf{Entail\_Explanation}: To go through the mill in the context here refers that vanity publishing will abuse and treat the work of literature being very poorly.
    \item[] \textbf{Contra\_Explanation}: Usually when we say go through the mill it does not mean something being celebrated and treated well but instead being abused and treated poorly which is being said here in the context of vanity publishing doing to the work of literature..
    \\
    \dots

\begin{table*}[t]
\renewcommand{\arraystretch}{1.2}
\begin{tabular}{ll}
\hline
Premise    & \begin{tabular}[c]{@{}l@{}}My father continuously gambles all his earnings away even though he has two minors to\\ support.\end{tabular}                                                                                                                             \\\hline
Hypothesis & \begin{tabular}[c]{@{}l@{}}My father is the best dad in the world who gambles all his earnings away and leaves me \\ and my sister struggling.\end{tabular}                                                                                                          \\\hline
Gold       & \begin{tabular}[c]{@{}l@{}}A great parent is someone who provides for their children and protects them but gambling \\ away all the money earned leaves the children in a difficult position and hence the father \\ cannot be considered the best dad.\end{tabular} \\\hline
FLUTE      & \begin{tabular}[c]{@{}l@{}}Gambling away one's earnings is not a good thing and it can lead to the father not being \\ able to support his children who are dependent on him.\end{tabular}                                                                           \\\hline
e-SNLI     & Gambling all his earnings away is not the best dad in the world.                                                                                                                                                                                                     \\\\\hline\hline
Premise    & He ran slowly.                                                                                                                                                                                                                                                       \\
Hypothesis & He ran like an olympic sprinter.                                                                                                                                                                                                                                     \\\hline
Gold       & \begin{tabular}[c]{@{}l@{}}An Olympic sprinter is someone who runs extremely fast, so saying someone ran like\\  one would imply they ran quickly, not slowly\end{tabular}                                                                                           \\\hline
FLUTE      & \begin{tabular}[c]{@{}l@{}}Olympic sprinters are known to run very fast, so saying someone ran like an olympic\\ sprinter would mean they ran very quickly\end{tabular}                                                                                              \\\hline
e-SNLI     & Slowly and sprinter are not the same                                                                                                                                                                                                                                 \\\\\hline\hline
Premise    & \begin{tabular}[c]{@{}l@{}}He's totally trustworthy as an executive secretary, but he's relatively new and \\ inexperienced and does not know what to do on occasions like this\end{tabular}                                                                         \\\hline
Hypothesis & \begin{tabular}[c]{@{}l@{}}He's totally trustworthy as an executive secretary, and has been around long enough\\ to know the ropes on occasions like this.\end{tabular}                                                                                              \\\hline
Gold       & \begin{tabular}[c]{@{}l@{}}To know the ropes means to understand or be familiar with the details, but the person is \\ relatively new and inexperienced\end{tabular}                                                                                                 \\\hline
FLUTE      & \begin{tabular}[c]{@{}l@{}}To know the ropes means to have experience in the job, but in this context the person is \\ relatively new and inexperienced\end{tabular}                                                                                                 \\\hline
e-SNLI     & \begin{tabular}[c]{@{}l@{}}He's not been around long enough to know the ropes and he is relatively new\\  and inexperienced.\end{tabular}                                                                                                            
\end{tabular}
\caption{\label{examples}Example of Sarcasm, Simile and Idiom NLI with respective gold and generated explanations from T5 by finetuning on FLUTE and e-SNLI}
\end{table*}

\section{Details of Human evaluation} \label{appendix:humaneval_details}
We follow \citet{Kayser_2021_ICCV} in all the below human evaluation procedures. Refer to Figure \ref{fig:mturk} for the example of the interface for crowdworkers. We first ask the crowdworkers to identify the relationship between the literal and figurative sentence (whether it is a contradiction or entailment). Using in-browser checks, we ensure that the crowdworkers understood the NLI pair by only accepting the submission if the relationship was identified correctly. 

Then, we provide 2 explanations: one generated by T5$_{\flute}$ and one generated by T5$_{\text{e-SNLI}}$. The crowdworkers do not know which one is which. For each NLE, we ask: \textit{Given the two sentences, does the explanation justify the answer above?}, and provide four options: Yes, Weak Yes, Weak No, and No. We also ask to provide the shortcomings of the explanations if the worker selected a score lower than Yes. The workers have the following options to choose from, following prior work by \cite{majumder2021rationale, Kayser_2021_ICCV}: \emph{Violates Common Sense}, \emph{Insufficient Justification}, \emph{Untrue to Input}, \emph{Too Trivial}, \emph{To Verbose}. Some examples of these shortcoming are provided in Table \ref{table:shortcomings_ex}. In addition to the Figure \ref{fig:shortcomings_bar}, we provide the bar plot of the number of shortcomings as percentage of the sample by figurative language type in Figure \ref{fig:shortcoms_by_figtype}.

We map the answers to $1, \frac{2}{3}, \frac{1}{3}, 0$ respectively. Then, we compute the average score across 3 workers per entry, and the sample average per figurative language type for the corresponding model.

\begin{figure*}[t]
    \centering
    \includegraphics[width=\textwidth]{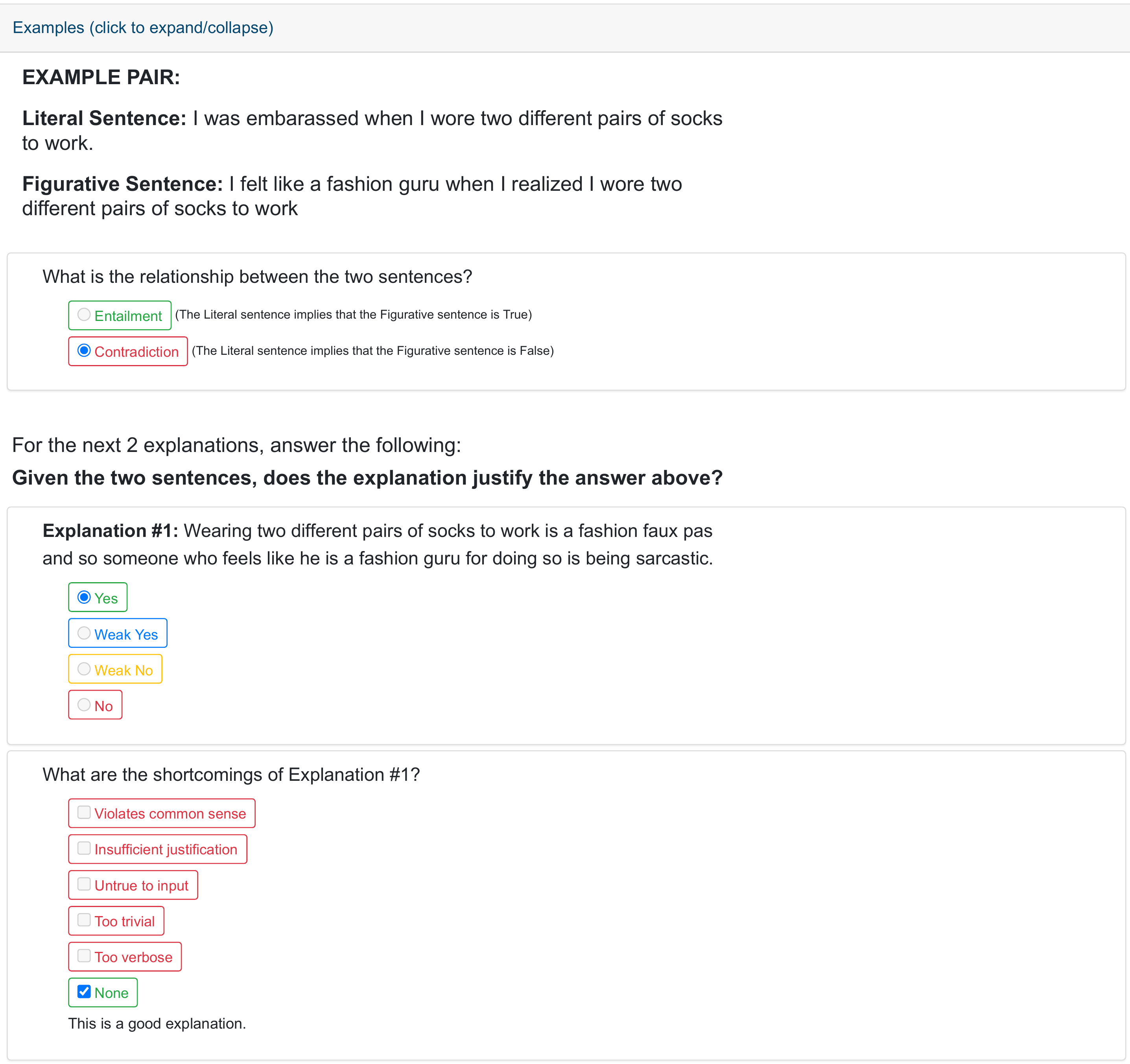}
    \caption{\label{mturk_expl} Amazon Mechanical Turk interface, borrowed from \cite{Kayser_2021_ICCV}, asking to first select the relationship between the sentences, and then evaluate the explanation.}
    \label{fig:mturk}
\end{figure*}

\end{itemize}

\begin{table*}[h]
\renewcommand{\arraystretch}{1.15}
\centering
\centering
\resizebox{\textwidth}{!}{%
\begin{tabular}{lllcll}
\hline
\multirow{2}{*}{\textbf{Shortcoming}}                                         & \multirow{2}{*}{\textbf{Premise}}                                                                                                                             & \multirow{2}{*}{\textbf{Hypothesis}}                                                                                                                                                                 & \multicolumn{1}{l}{\multirow{2}{*}{\textbf{Label}}} & \multicolumn{2}{c}{\textbf{Explanation}}                                                                                                                                                                                                                                                                         \\ \cline{5-6} 
                                                                              &                                                                                                                                                               &                                                                                                                                                                                                      & \multicolumn{1}{l}{}                                & \multicolumn{1}{c}{\textbf{T5$_{\text{e-SNLI}}$}}                                                                               & \multicolumn{1}{c}{\textbf{T5$_{\flute}$}}                                                                                                                                                    \\ \hline
\textbf{\begin{tabular}[c]{@{}l@{}}Violates\\ Common\\ Sense\end{tabular}}    & The gesture was awkward.                                                                                                                                      & \begin{tabular}[c]{@{}l@{}}The gesture was\\ like a ballet move\end{tabular}                                                                                                                         & C                                                   & \begin{tabular}[c]{@{}l@{}}A gesture is not \\ a ballet move.\end{tabular}                                           & \begin{tabular}[c]{@{}l@{}}A ballet move is \\ graceful and fluid, \\ while an awkward \\ gesture is not \\ graceful or fluid.\end{tabular}                                               \\ \hline
\textbf{\begin{tabular}[c]{@{}l@{}}Insufficient\\ Justification\end{tabular}} & \begin{tabular}[c]{@{}l@{}}I stayed in a house with\\  a rat problem for a while, \\ it was not fun.\end{tabular}                                             & \begin{tabular}[c]{@{}l@{}}I simply cannot \\ wait to stay in \\ a house again \\ that had a rat \\ problem because \\ it was not only\\ a lot of fun but also \\ super enjoyable\end{tabular}       & C                                                   & \begin{tabular}[c]{@{}l@{}}It was either not \\ fun or not fun.\end{tabular}                                         & \begin{tabular}[c]{@{}l@{}}Rats are often \\ associated with \\ being filthy and \\ unhygienic and \\ so staying in a house \\ with rat problems \\ would not be fun at all.\end{tabular} \\ \hline
\textbf{\begin{tabular}[c]{@{}l@{}}Untrue to \\ Input\end{tabular}}           & \begin{tabular}[c]{@{}l@{}}Even though he was \\ strong in his magic, \\ it felt weak compared \\ to the inferno I'd faced \\ when I killed Mab.\end{tabular} & \begin{tabular}[c]{@{}l@{}}He was strong \\ in his magic but, \\ compared to the \\ blazing inferno that \\ i'd faced when i'd \\ killed mab, his \\ power felt like \\ a candle flame.\end{tabular} & E                                                   & \begin{tabular}[c]{@{}l@{}}The candle flame \\ is the inferno.\end{tabular}                                          & \begin{tabular}[c]{@{}l@{}}A candle flame is a \\ very small flame, \\ which is weak compared \\ to blazing inferno.\end{tabular}                                                         \\ \hline
\textbf{Too Trivial}                                                          & He ran fast.                                                                                                                                                  & \begin{tabular}[c]{@{}l@{}}He ran like \\ an olympic sprinter.\end{tabular}                                                                                                                          & E                                                   & \begin{tabular}[c]{@{}l@{}}He ran fast like \\ an olympic sprinter.\end{tabular}                                     & \begin{tabular}[c]{@{}l@{}}An olympic sprinter \\ is a very fast person, \\ so if he ran like an \\ sprinter, it would mean \\ that he ran fast.\end{tabular}                             \\ \hline
\textbf{Too Verbose}                                                          & \begin{tabular}[c]{@{}l@{}}She asked the witnesses \\ for information.\end{tabular}                                                                           & \begin{tabular}[c]{@{}l@{}}She petitioned the \\ witnesses for \\ information.\end{tabular}                                                                                                          & E                                                   & \begin{tabular}[c]{@{}l@{}}She asked the \\ witnesses for \\ information and \\ petitioned the witness.\end{tabular} & \begin{tabular}[c]{@{}l@{}}Petitioning someone \\ means asking them for \\ information, which is \\ what she did \\ in this sentence.\end{tabular}                                        \\ \hline
\end{tabular}%
}
\caption{\label{table:shortcomings_ex} Examples of shortcomings of T5$_{\text{e-SNLI}}$ explanations. For this table, T5$_{\text{e-SNLI}}$ explanations were sorted by most crowd workers votes for a respective shortcoming.}
\end{table*}

\begin{figure*}[t]
    \centering
    \includegraphics[width=\textwidth]{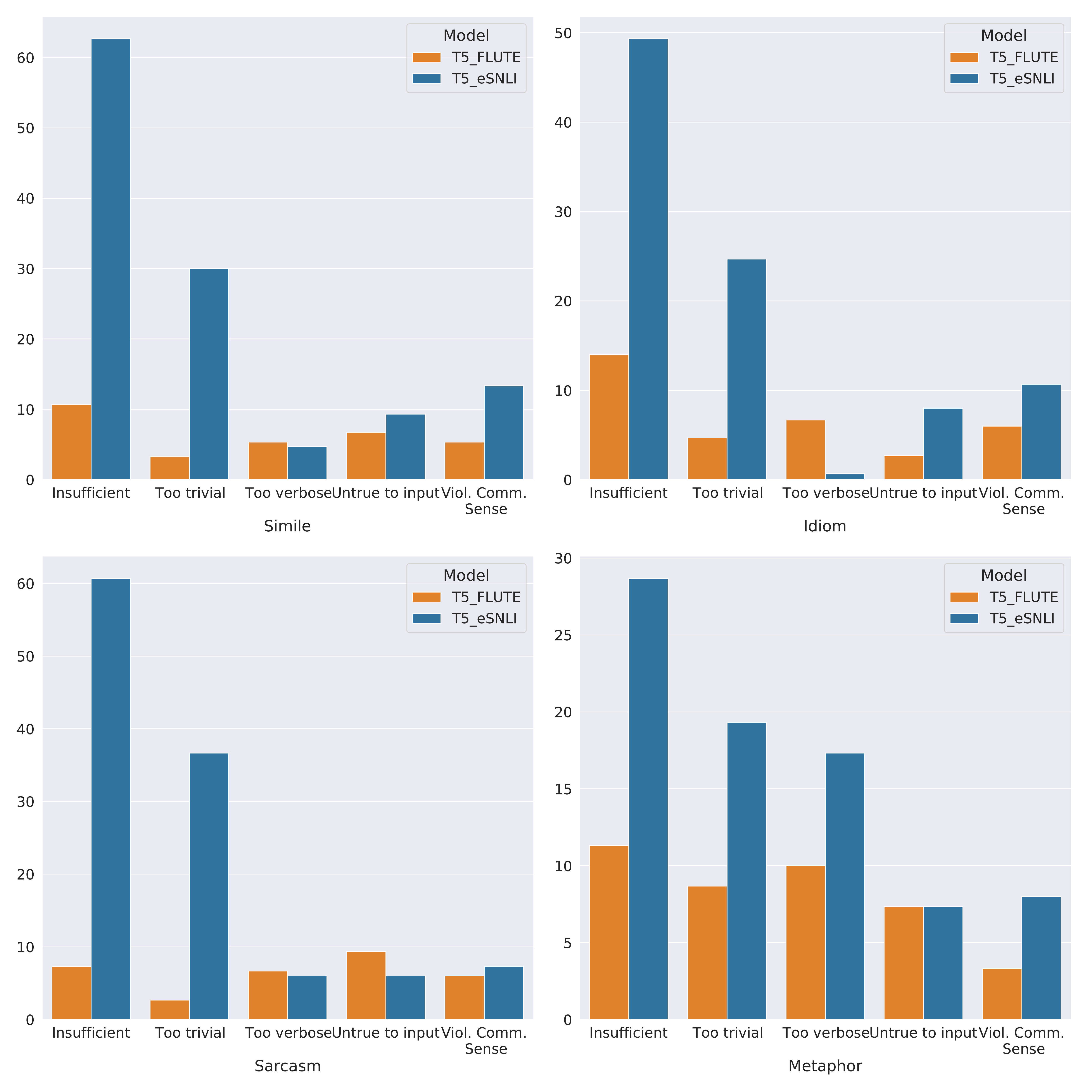}
    \caption{Bar plot of the number of crowd worker-identified shortcomings of explanations generated by T5$_{\text{e-SNLI}}$ and T5$_{\flute}$ by type of shortcoming, figurative language type, and by type of model as percent of the sample (lower means fewer shortcomings). }
    \label{fig:shortcoms_by_figtype}
\end{figure*}

\end{document}